\documentclass[letterpaper, 10 pt, conference]{ieeeconf}  
\IEEEoverridecommandlockouts                              

\title{\LARGE \bf
Environmental Sampling with the Boustrophedon Decomposition Algorithm
}

\author{Hannah He\textsuperscript{1}, Joe Norby\textsuperscript{2}, Sean Wang\textsuperscript{2}, Natasha Sihota\textsuperscript{3},\\
Thomas P. Hoelen\textsuperscript{3}, Gregory V. Lowry\textsuperscript{4}, and Aaron M. Johnson\textsuperscript{2}\\[.5em]
{\normalsize \textsuperscript{1}Computer Science, \textsuperscript{2}Mechanical Engineering, \textsuperscript{3}Chevron CTC, \textsuperscript{4}Civil and Environmental Engineering}\\
{\normalsize Carnegie Mellon University, Pittsburgh, PA, USA}\\
{\normalsize \{hannahhe, jnorby, sjw2, glowry, amj1\}@andrew.cmu.edu }%
}
\date{\vspace{-1em}}

\usepackage{graphicx}
\usepackage{amsmath}
\usepackage{amssymb}
\usepackage{amsfonts}
\usepackage{caption}
\usepackage{subcaption}
\usepackage{float}
\usepackage{cite}

\usepackage{epsfig} 
\usepackage{times}
\usepackage{mathptmx}

\usepackage[bookmarks=true]{hyperref}
\hypersetup{%
  breaklinks,
  bookmarksnumbered=true,
  bookmarksopen=true,
  colorlinks=true,
  linkcolor=black,
  urlcolor=black,
  citecolor=black
}

\begin{document}

\maketitle

\begin{abstract}
The automation of data collection via mobile robots holds promise for increasing the efficacy of environmental investigations, but requires the system to autonomously determine how to sample the environment while avoiding obstacles. Existing methods such as the boustrophedon decomposition algorithm enable complete coverage of the environment to a specified resolution, yet in many cases sampling at the resolution of the distribution would yield long paths with an infeasible number of measurements. Downsampling these paths can result in feasible plans at the expense of distribution estimation accuracy. This work explores this tradeoff between distribution accuracy and path length for the boustrophedon decomposition algorithm. We quantify algorithm performance by computing metrics for accuracy and path length in a Monte-Carlo simulation across a distribution of environments. We highlight conditions where one objective should be prioritized over the other and propose a modification to the algorithm to improve its effectiveness by sampling more uniformly. These results demonstrate how intelligent deployment of the boustrophedon algorithm can effectively guide autonomous environmental sampling.
\end{abstract}

\section{Introduction}
Environmental sampling is the process of extracting information from a given environment by collecting measurements at different locations and analyzing the data. For example, environmental sampling has been used for mineral prospecting \cite{LIMA20031853}, characterization of algae blooms \cite{4118466}, and air particle monitoring \cite{doi:10.1080/02786826.2019.1623863}. Techniques for environmental sampling can also be deployed to aid environmental remediation, as knowledge of underlying contamination distributions can more efficiently focus remediation efforts.

Characterizing contamination within an environment is typically a costly and time-consuming process which relies on manual sample extraction and laboratory post-processing \cite{cline1944principles}. These dependencies render widespread and dense sampling impractical, which limits the accuracy of environmental investigations. Recent advances in mobile robotic technology offer promising solutions to these problems by automating the sampling process \cite{dunbabin2012robots}. Field sensors to detect contaminant concentration levels mounted on mobile robotic platforms could take measurements \textit{in situ}, greatly increasing the number of samples the system could gather and thus increase the fidelity of the study.

\begin{figure}[t]
    \centering
    \includegraphics[height = 1.8in]{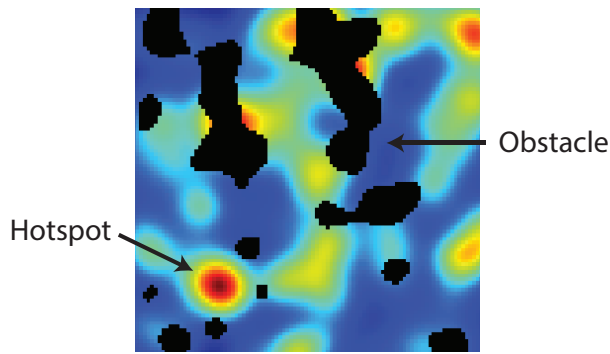}
    \caption{An example environment for autonomous sampling. The underlying distribution is shown as a color map with warmer colors indicating higher concentration levels, and obstacles to robot motion are shown in black.}
    \label{fig:sampleEnv}
\end{figure}

One challenge of autonomous environmental sampling is implementing an appropriate search algorithm. The search algorithm takes prior knowledge about the environment and decides where to take measurements and how to travel between sample locations. These algorithms must be able to appropriately cover the area of interest with measurements to estimate the underlying contaminant distribution or locate hotspot regions of high levels of contamination. They must also ensure that the robot is able to feasibly traverse the resulting path, and therefore must reason about obstacles to robot navigation or areas that may involve more risk to traverse (e.g. uneven terrain or regions with uncertain features). An example of such a region with an underlying contaminant distribution is shown in Fig.~\ref{fig:sampleEnv}.

This class of planning problems is known as \textit{coverage planning}, and is motivated by similar applications such as lawn mowing, vacuum cleaning, and farming. Although this problem is NP-hard, many approximate methods exist for a number of applications \cite{lavalle2006planning}. One simple yet effective approach for solving this problem is the boustrophedon decomposition algorithm \cite{choset1998coverage}. This method drives the robot in a rastering motion, which is similar to the classic back and forth motion of a lawn mower. Repeating this pattern results in a full coverage path that guarantees any location in the freespace (the space of all valid robot configurations that do not intersect an obstacle) is within a certain distance of a sample location. This algorithm can be extended to handle obstacles by segmenting the environment into a set of obstacle-free regions known as cells which can be rastered independently.

However, one primary difference between environmental sampling and these other coverage applications is that environment samples are discretely drawn from a continuous distribution, rendering complete coverage impossible. Naively implementing boustrophedon decomposition with high sample density may result in excessively long periods of data collection, or miss key contaminant information if sample density is drastically reduced. A good search algorithm should therefore maximize the information gained through the environmental sampling process while minimizing the operational costs. In particular, the resulting estimated distribution should closely represent the actual distribution in the environment and encapsulate important features such as the location of hot spots. While meeting these requirements, the amount of time and distance traveled to measure all sampling locations should be minimized and the path through all sampling locations should satisfy traversability requirements. Several prior works have applied boustrophedon decomposition to environmental monitoring applications and investigated near-optimal methods of decomposing the environment and selecting rastering schemes within cells for a fixed sample density, but have not directly considered or quantified the effects of sample density on performance or modified the algorithm to properly maintain sample spacing \cite{karapetyan2018multi,karapetyan2019riverine,pham2017aerial}.


This report evaluates the performance of the boustrophedon decomposition algorithm for environmental sampling applications and presents modifications to the algorithm to improve its performance in this domain. For performance evaluation, we implement the algorithm inside a Monte-Carlo framework and measure the expected performance of different performance metrics across randomized environment features. Section \ref{sec:boustrophedon} describes the general boustrophedon algorithm in greater detail. Section \ref{sec:performanceEvalFramework} describes how the algorithm's performance was evaluated. Section \ref{sec:results} presents analysis of the results of the performance evaluation experiment. Lastly, Section \ref{sec:preprocessing} describes improvements made to the shorten the path length of the original boustrophedon algorithm and quantifies their effects on algorithm performance.

\section{Boustrophedon Decomposition Algorithm}
\label{sec:boustrophedon}
This section provides an overview of the boustrophedon decomposition algorithm. The algorithmic pipeline takes two inputs: a map of the environment and a desired sample spacing distance. The environment map is encoded as a 2D occupancy grid, where a `1' represents a obstacle to the robot's navigation (such as a tree or a building), and a `0' represents traversable terrain.  The desired sample spacing controls the distance (measured in units of grid cells) between measurements and therefore tunes the density of sampling as well as the width of the raster. After the robot moves a number of cells equal to the sample spacing, the robot takes a measurement at that location and continues rastering, with the exception of traveling between cells (described below).
The resulting path a robot would take through this environment and the areas where the robot sampled are the outputs of this pipeline. 

\subsection{Cellular Decomposition}
\begin{figure}[t]
    \centering
    \includegraphics[height=3in]{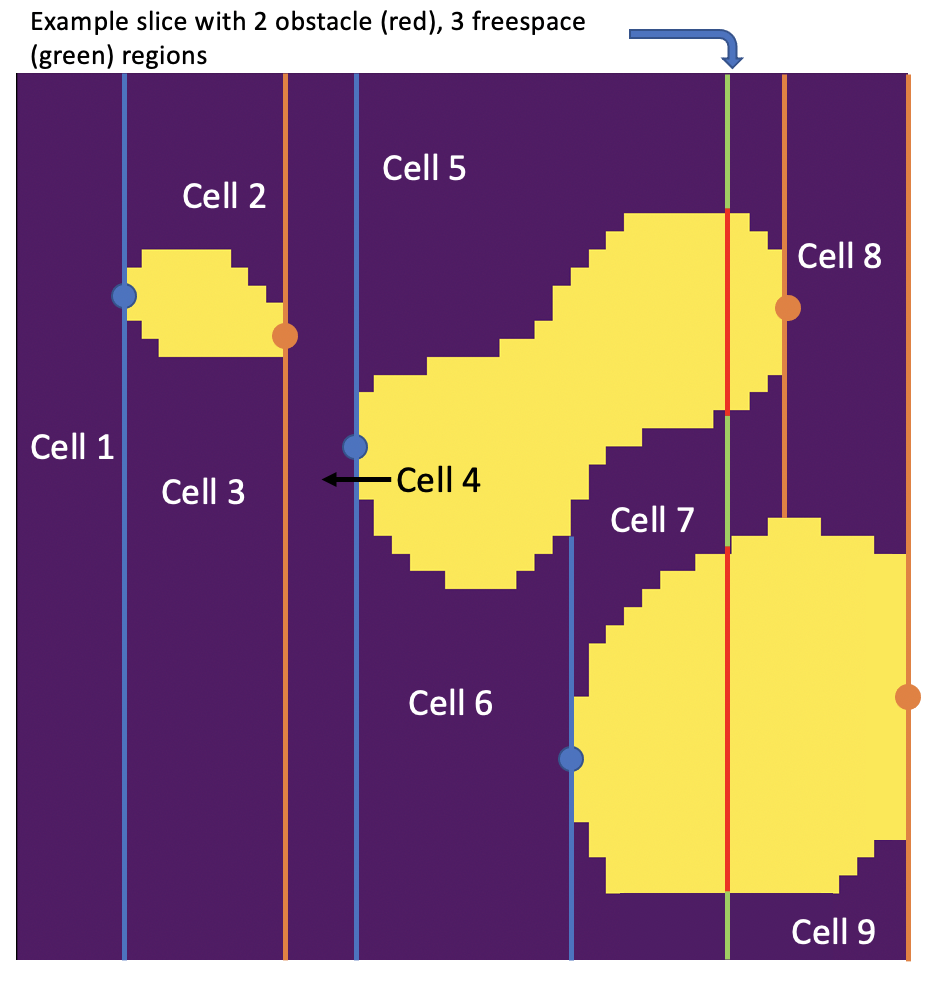}
    \caption{Example of boustrophedon decomposition process. Obstacles are shown in yellow. The dotted line shows an example of a vertical slice where three freespace regions are counted. Blue dots represent IN events. The blue lines from the IN events close off a cell and open two new cells. The orange dots represent OUT events. The orange lines from OUT events close off two cells and open one new cell.}
    \label{fig:example_decomposition}
\end{figure}

The boustrophedon decomposition algorithm starts by decomposing the environment into smaller, obstacle-free regions called cells through a flood fill approach. This is done by looking through individual vertical slices through the environment. For each slice, the number of connected freespace regions are counted. For example, the vertical slice shown as the red and green line in Figure \ref{fig:example_decomposition} contains three freespace regions and two obstacle regions. Going left to right, an increase of freespace regions in the vertical slices are considered as an IN event. A decrease of freespace regions is considered as an OUT event. For IN events, the cell covering the location where the IN event occurred is closed and two new cells are formed. For example, in Figure \ref{fig:example_decomposition}, the leftmost blue dot represents an IN event. Since Cell 1 is covering the IN event location, it is closed and two new cells are formed (Cell 2 and Cell 3). If the location of the IN event is not covered by a cell, then no cells are closed, but one new cell is formed. For example, since the IN event represented by the rightmost blue dot is not covered by any cell, no cells are closed, and two new cells are formed (Cell 7 and Cell 9). For an OUT event, the two cells neighboring the OUT event locations are closed and a new cell is formed. For example, at the OUT event represented by the left most orange dot, Cell 2 and Cell 3 are closed and Cell 4 is formed. This process is iterated until the whole environment has been split into individual cells. Each cell can then be completely rastered by the robot without hitting any obstacles.

\subsection{Intracellular Rastering}
Once the environment is broken up into smaller cells, the robot can raster each individual cell so that collectively, after the robot has rastered all of the cells, the entire environment will be traversed. The algorithm assumes that the robot will always start in one of the four corners of a cell. The direction of rastering (up or down and left or right) is determined based on the start position, or whichever corner of the cell the robot starts in. For instance, if the robot starts in the upper left corner of a cell, then it will start traveling downwards and move horizontally to the right until it reaches the right hand border of the cell. Since each cell is represented by the ceiling and floor coordinates that border it, the robot simply has to travel up or down until it hits a coordinate in the ceiling or floor (or, equivalently, until it reaches an obstacle in the map). It then moves to the right or left in the ceiling/floor coordinates by the amount specified by the path width (or the length of the cell if the path width is larger than the cell width) before continuing to travel in the opposite direction as it was before. The entire cell is guaranteed to be safely and completely rasterable because each cell is represented by the traversable coordinates bordering it, and if an obstacle borders the cell, the closest a robot is able to get to an obstacle is the coordinate before the obstacle.

\subsection{Intercellular Travel}
After a cell has been successfully rastered, the algorithm must decide which cell to raster next. Each cell has four corner points at which the robot can start rastering. In our implementation, the robot greedily travels to the closest corner of unrastered cells and rasters the corresponding cell. In the future, the robot could lower its overall path length by using an A* algorithm to optimize the order of cells to raster and the entry and exit points of each cell.

\subsection{Overall Path}
\begin{figure*}
\centering
\begin{subfigure}{0.45\linewidth} 
  \centering
  \includegraphics[width=1.0\linewidth]{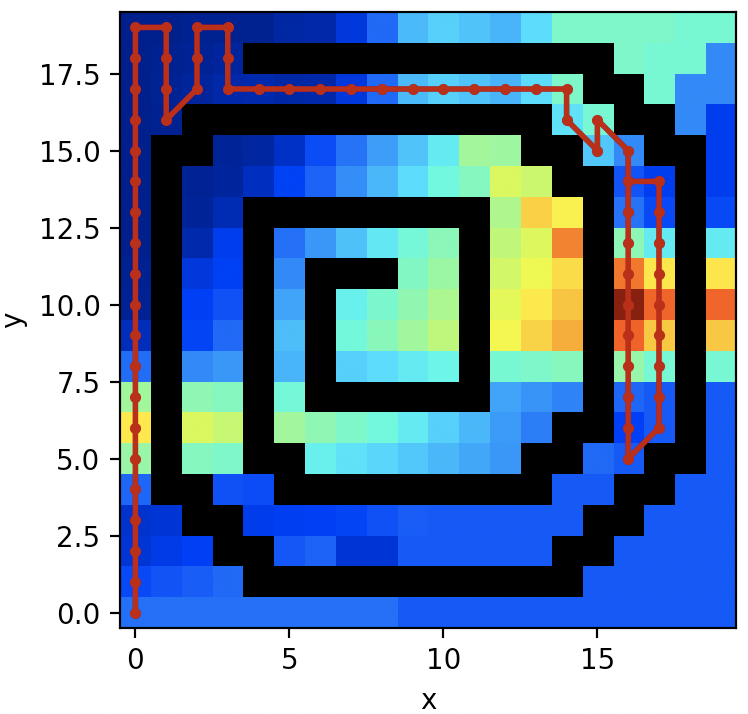}
  \caption{Solving a spiral map without boustrophedon cellular decomposition}
  \label{fig:BDPathExample_no_decomp}
\end{subfigure}
\hfil
\begin{subfigure}{0.45\linewidth} 
  \centering
  \includegraphics[width=1.0\linewidth]{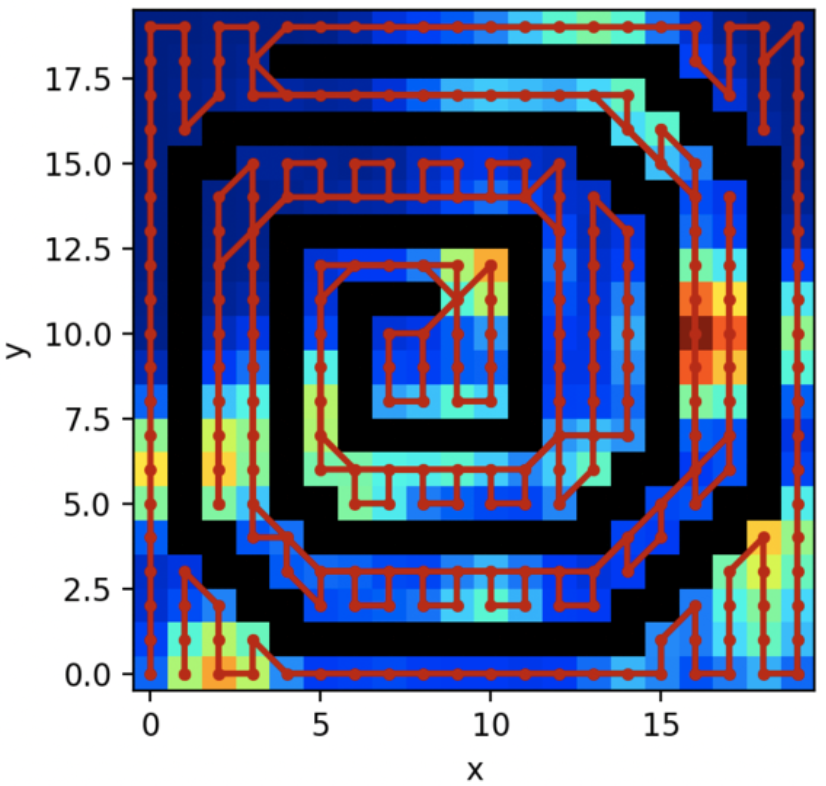}
  \caption{Solving a spiral map with boustrophedon cellular decomposition}
  \label{fig:BDPathExample_with_decomp}
\end{subfigure}
\caption{Using the boustrophedon cellular decomposition method allows for coverage solutions in obstacle filled environments that would not be possible with simple boustrophedon. Figure~\ref{fig:BDPathExample_no_decomp} shows that the robot gets stuck in the spiral map if it does not first break apart the environment into cells. In contrast, Figure~\ref{fig:BDPathExample_with_decomp} shows that once the environment is decomposed into traversable cells, the robot can simply raster each cell sequentially to guarantee coverage. In each figure, obstacles are overlaid in black and sample points are shown as red dots. Red lines indicate the sampling path.}
\label{fig:BDPathExample}
\end{figure*}

The above steps of rastering a cell and determining which cell to raster next is repeated until there are no more reachable, unrastered cells in the environment, therefore fully covering the freespace. The entire robot path, including the raster path of each cell and the path traveled between each cell as well as the list of sampled coordinates are condensed into two arrays of tuples (coordinates) and returned. An example of the path traveled using the boustrophedon decomposition algorithm is shown in Figure \ref{fig:BDPathExample}.

\section{Performance Evaluation Framework}
\label{sec:performanceEvalFramework}
\begin{figure*}[t]
    \centering
    \includegraphics[width=\linewidth]{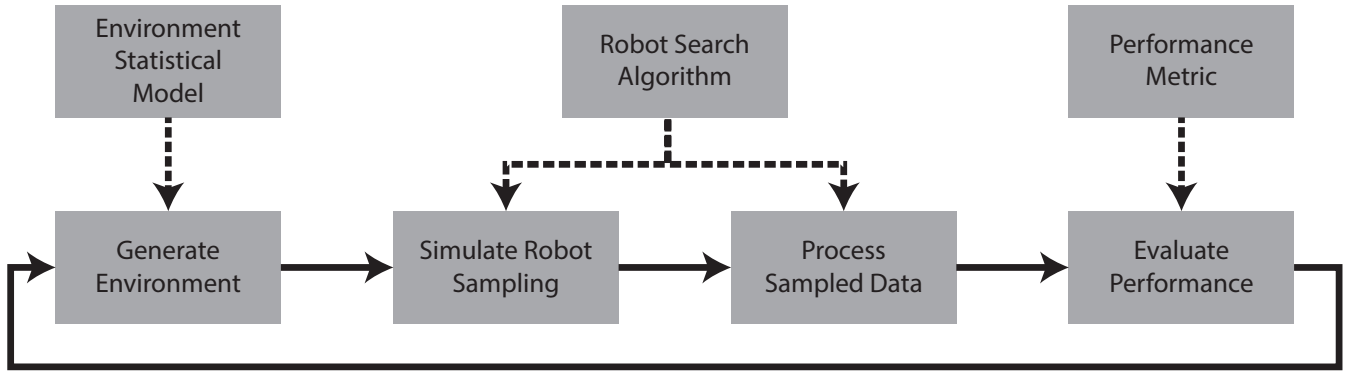}
    \caption{Diagram of Monte Carlo performance evaluation framework.}
    \label{fig:monte_carlo}
\end{figure*}

This section discusses the framework created to evaluate the environmental sampling performance of the boustrophedon and other planning algorithms. This framework relies on Monte Carlo methods to quantify this performance over a large set of randomized environments. The primary benefit of this analysis is its ability to efficiently explore relationships between any number of different planning algorithms, sets of environments, and their underlying parameters. A brief summary of Monte Carlo methodology is given in Section~\ref{sec:monte_carlo}, followed by a description in Section~\ref{sec:environment_generation} of the randomized environment generation employed by this framework. Lastly, in Section~\ref{sec:performance_evaluation_methodology} we present the methodology that quantifies the performance of a planning algorithm. A diagram depicting the performance evaluation framework is shown in Figure \ref{fig:monte_carlo}. This methodology is demonstrated in Section~\ref{sec:results} with results for the boustrophedon algorithm.

\subsection{Summary of Monte Carlo Methods} \label{sec:monte_carlo}
Monte Carlo simulations are useful tools for computing the expected value of complex functions with inputs randomly sampled from some probability distribution. In our case, we would like to measure the expected value of the search algorithm's performance given environments that are randomly sampled from a probability distribution. That is to say, we would like to compute:
\begin{equation}
    l = \mathbb{E}[H(x)] = \int H(x)f(x)dx,
\end{equation}
where $x$ is a random variable that represents an environment, $f$ is a probability density function of environments, and $H$ is a function that evaluates the performance of the search algorithm given a particular environment. Since the function $H$ is quite complex calculating the integral in closed form is not possible.

Instead of calculating the expected value $l$ in closed form, a Monte-Carlo method aims to numerically approximate the expected value with the following equation.
\begin{equation}
    \hat{l} = \frac{1}{N} \sum_{k=1}^{N}H(x_k) \approx \mathbb{E}[H(x)] \qquad x_1,...,x_N \sim f.
\end{equation}
In this process, $N$ environments ($x_1,...,x_N)$ are sampled from the probability distribution of environments. Then the performance metric function $H$ is evaluated for each sampled environment. The mean of the resultant values is then taken as an approximation of the expected performance of the search algorithm. To get an accurate approximation of $l$, $N$ should be large to provide a set of samplings that closely represents the probability distribution function $f$.

The process for sampling an environment from the probability distribution of environments $f$ is described in Section \ref{sec:environment_generation}. Different performance metric functions $H$ are described in Section \ref{sec:performance_evaluation_methodology}.

\subsection{Environment Generation} \label{sec:environment_generation}

\begin{figure*}[t]
  \centering
  \includegraphics[width=1.0\linewidth]{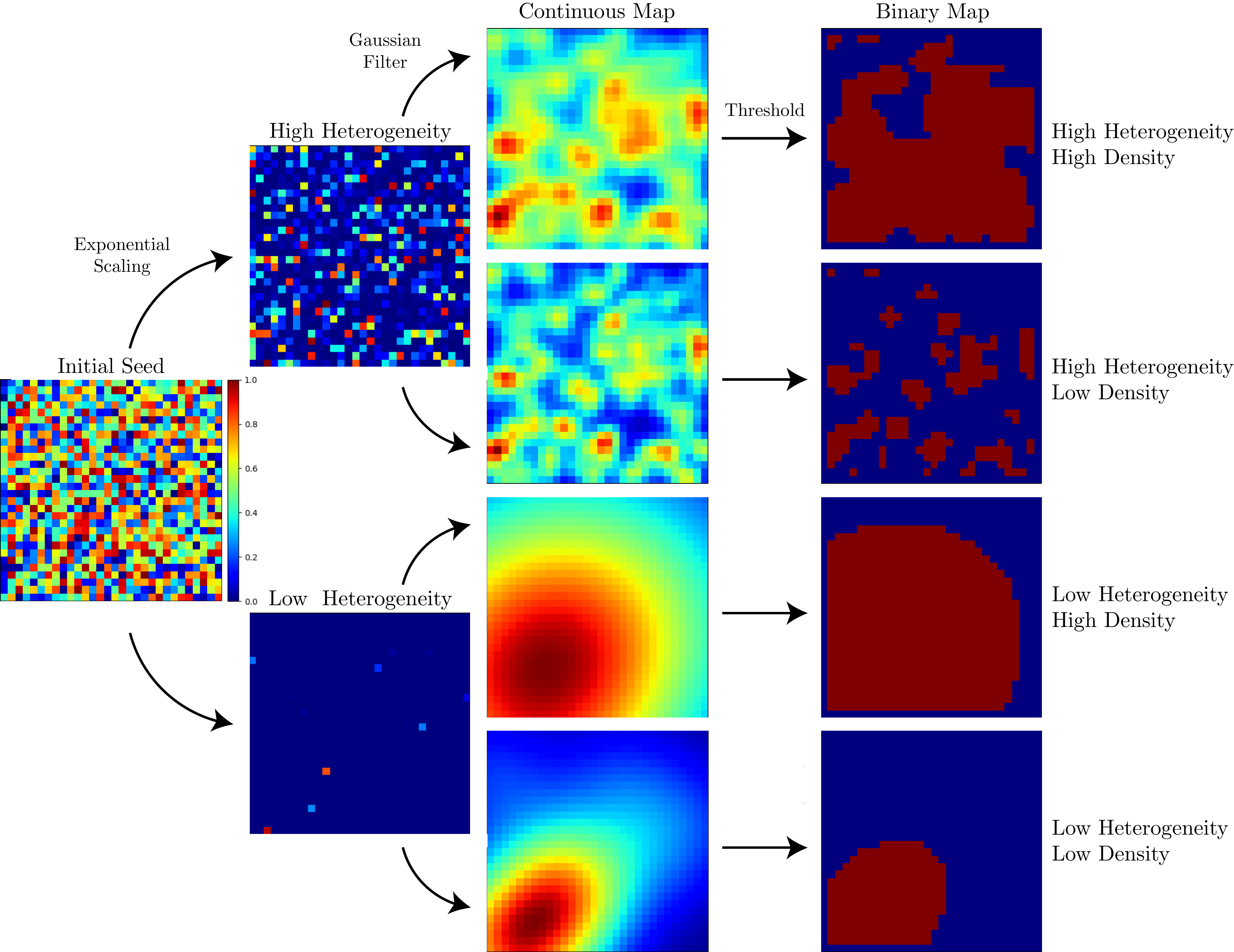}
  \caption{Both analyte distribution and obstacle maps are randomly generated, parameterized by density (the percentage of area covered by hotspots or obstacles) and heterogeneity (how spread out hotspots or obstacles are). All maps start with a random initial seed of values between zero and one, then are transformed by the heterogeneity and density parameters to represent environments of interest.}
  \label{fig:environment_generation}
\end{figure*}

Each environment consists of a smooth distribution (which can be thresholded to yield the underlying binary hotspot map) and a binary obstacle map. Both the distribution and obstacle map are generated randomly from a small set of parameters that describe their physical properties. This process is illustrated in Figure~\ref{fig:environment_generation}. We have selected two parameters in particular to describe the properties of a map: density and heterogeneity. Density measures how much of the map is covered by either hotspots or obstacles, and heterogeneity measures the spread of these features. First, each pixel $i$ in the map is seeded with a random number $s_i \in [0,1]$. The image is then scaled exponentially by the heterogeneity parameter to obtain varying spreads of hotspots or obstacles with 
\begin{align}
    y_i = 10^{\left(-\frac{s_i}{h} \right)}
\end{align}
where $y_i$ is the scaled value of the map at location $i$ and $h$ is the heterogeneity parameter. This process introduces the notion of hotspots, and is equivalent to sampling from an exponential distribution parameterized by $h$. A Gaussian filter then spatially smooths these maps to obtain the desired density, replicating analyte or obstacle dispersion. Lastly, a threshold is applied to yield a corresponding binary map. This is equivalent to finding the filter variance $\sigma$ that satisfies
\begin{align}
    y_i &= G_i(x,\sigma) \\
    \rho_{\text{des}} &= \frac{1}{M}\sum_{i=1}^M u(y_i - \epsilon)
\end{align}
where $\rho_{\text{des}}$ is the desired density, $M$ is the number of locations in the map, $G_i(x,\sigma)$ is the resulting value at location $i$ after applying a Gaussian filter to environment $x$ with variance $\sigma$, $u(\cdot)$ is the Heaviside step function, and $\epsilon$ is the value of the threshold. Hotspot and obstacle maps use this binary representation, while the analyte distribution uses the smooth counterpart of the hotspot map.
Figure~\ref{fig:environment_generation} also shows the effect each of these parameters have on the qualitative appearance of the environments, showing the range of environments that can be generated with this method. For example, environments with a few large obstacles (such as flat areas with buildings) can be represented with medium heterogeneity and high density, while environments with very sparse obstacles (such as small clusters of trees) can be represented with high heterogeneity and low density.

The density and heterogeneity parameters are uniformly sampled to generate random environments, but it may be the case that in the future different map metrics would be desirable to investigate such as average feature size, number of features, average distance to the closest feature, and so on. Monte Carlo methods are well suited to comparing across other metrics. As long as one can numerically approximate the probability that an environment with a given metric will be calculated, comparisons with that metric can be made. Additionally, this same method could be made to more accurately reflect performance in a desired physical location by determining those probability functions for candidate locations.

\subsection{Performance Evaluation Methodology} \label{sec:performance_evaluation_methodology}

The performance of a particular planning algorithm over a given environment is a function of two objectives: maximizing accuracy and minimizing effort. Clearly maximizing accuracy is desirable to provide more useful information about the environment. However since a sampling system is subject to constraints on energy consumption and operation time, short paths with a few efficient sample locations are desirable over long, inefficient ones.

To quantify this tradeoff, we select metrics to assess different aspects of performance. For a given parameter of interest (e.g.\ sample spacing, as discussed in Section~\ref{sec:results}), environments are randomly generated and the planning algorithm executed to yield the desired path and sample locations. The estimated distribution is then created by constructing a triangular mesh from the sampled locations, and linearly interpolating within each triangle. Estimates for points in the map outside the convex hull of the sampled locations (e.g.\ points on the edge of the map) are assigned the measurement taken from the nearest sample. 

We quantify the overall distribution recreation accuracy with root-mean-squared error (RMSE), defined by
\begin{align}
    \text{RMSE} = \sqrt{\frac{1}{M} \sum_{i=1}^M (y_i - \hat y_i)^2}
\end{align}
where $y_i$ is the true value of the distribution at location $i$ and $\hat y_i$ is value from the estimated distribution at that location. 

While the RMSE metric quantifies the overall accuracy of the estimated distribution, often this aggregate performance is less important than simply finding the locations where the concentration is greatest. The performance in finding these hotspots is quantified by the hotspot miss rate (HMR). This is calculated by iterating through each hotspot, recording whether any samples were taken within them, and dividing the number of hotspots with no samples by the total number.
\begin{align}
    \text{HMR} = \frac{\text{number of missed hotspots}}{\text{number of hotspots}}
\end{align}
In calculating both accuracy metrics, data covered by obstacles are not considered.

The other two metrics are path length and number of samples, each of which quantify the effort required to obtain the estimated distribution. Often these two correlate, although for applications where sampling time greatly differs from travel time their relative importance shifts.

In this report all measurements are assumed to be perfect, but this framework could easily accommodate uncertainty in sampling and calculate the resulting performance sensitivity. 

\section{Results} \label{sec:results}
\begin{figure*}[t]
\centering
\includegraphics[width=\linewidth]{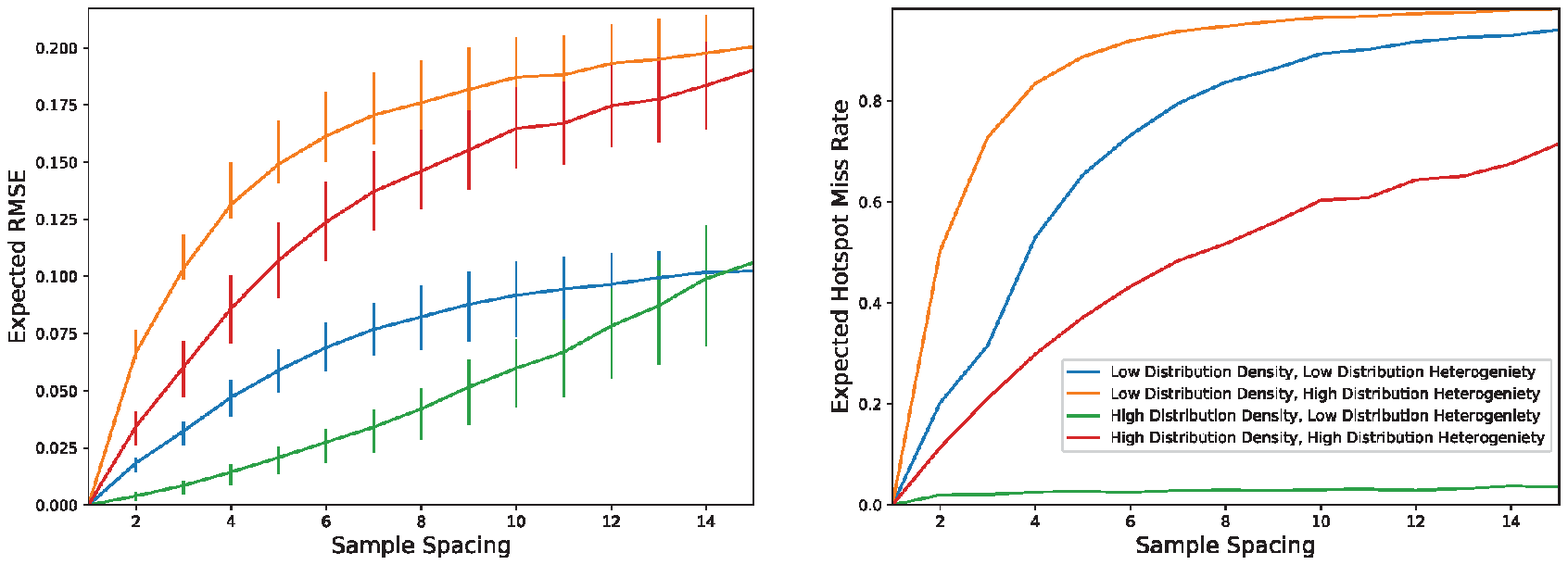}
\caption{Expected performance of boustrophedon decomposition computed using Monte-Carlo simulations. Performance is compared between different environment distributions and for plans with different sample spacings.}
\label{fig:performanceVspacing}
\end{figure*}

The expected values of the performance metrics, described in Section \ref{sec:performance_evaluation_methodology}, over the randomly generated terrain were evaluated for a variety of sample spacings. For each sample spacing, a Monte-Carlo simulation is run for 10,000 trials. Figure \ref{fig:performanceVspacing} shows the results for four different types of distributions: low or high density (0.01 or 0.5) and low or high heterogeneity (0.1 or 0.75). For all trials, low density (0.01) and low heterogeneity (0.1) obstacle maps were used.

\begin{figure*}[t]
\vspace{4em}
\begin{subfigure}{1.0\textwidth} 
  \centering
  \includegraphics[width=1.0\linewidth]{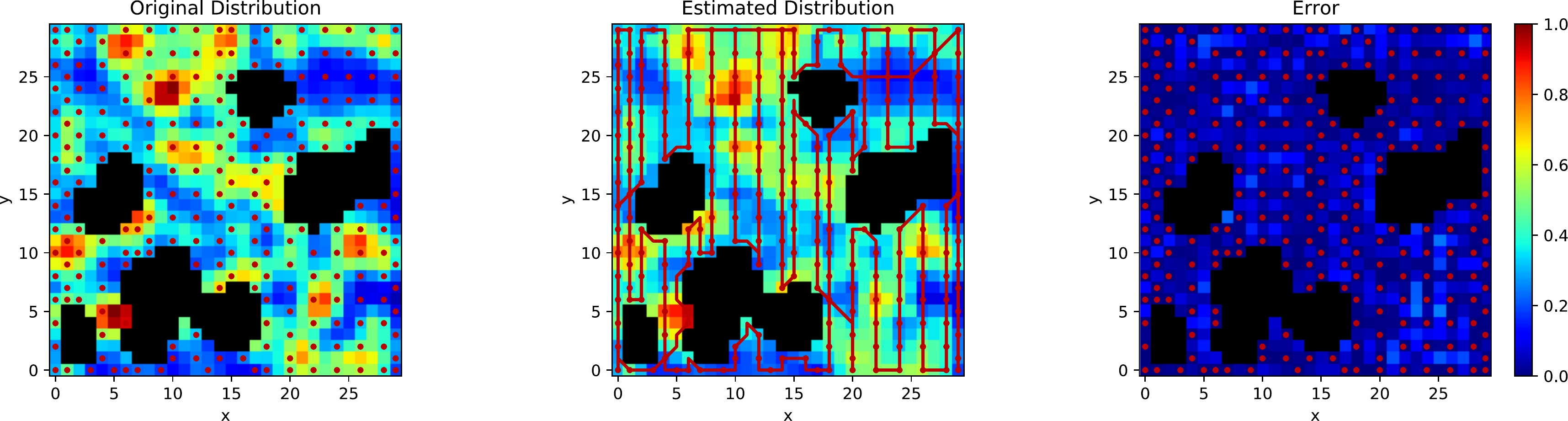}
  \caption{Small sample spacing}
\end{subfigure}
\vspace{5mm}
\newline
\begin{subfigure}{1.0\textwidth} 
  \centering
  \includegraphics[width=1.0\linewidth]{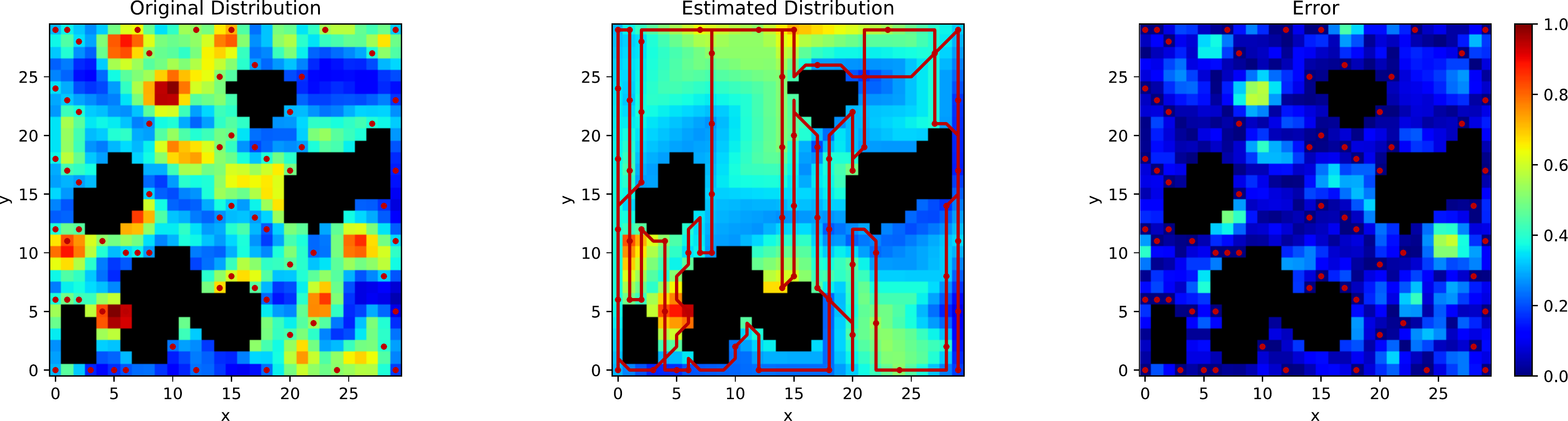}
  \caption{Large sample spacing}
\end{subfigure}
\caption{Sample spacing has a visible effect on estimation error. Coarse rastering misses many of the hotspots that finer rastering is able to detect, but the overall path length is shorter. The leftmost axes show the original contaminant distributions, the central axes show the estimated distributions, and the right shows the error between the two. In each figure, obstacles are overlaid in black and sample points are shown as red dots. Red lines indicate the sampling path.}
\label{fig:results_example}
\end{figure*}

Since boustrophedon decomposition is a complete coverage algorithm, using a sample spacing of 1 led to the robot sampling at every possible location in the environment resulting in no distribution recreation error, and no undetected hotspots. As the sample spacing increases, the recreation error increase and more hot spots were undetected. Environments with high heterogeneity distributions have less uniform distributions and thus require more samples and lower sample spacing to reduce overall error (RMSE). Envrionments with low distribution densities have smaller hotspots, and thus require lower sample spacing to reduce the number of undetected hotspots. Figure \ref{fig:results_example} shows and example environment where these effects are apparent.

\begin{figure*}[t]
\centering
\includegraphics[width=0.8\linewidth]{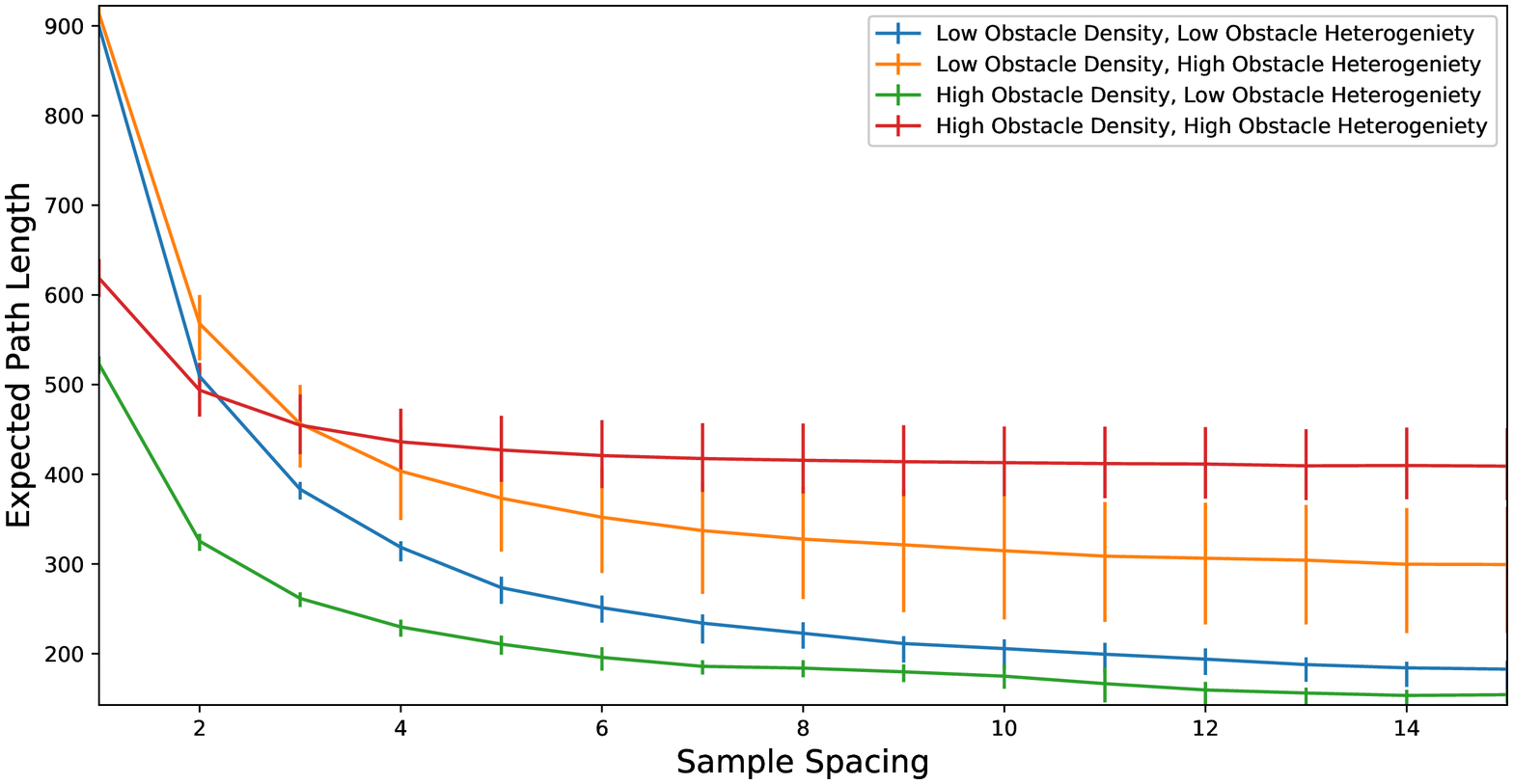}
\caption{The expected path length of the robot using boustrophedon cellular decomposition as the search algorithm with varying sample spacing is found by running a Monte-Carlo simulation on randomly generated 30 by 30 maps with four different obstacle types. The expected path length of the robot decreases with the sample spacing.}
\label{fig:pathLengthVspacing}
\end{figure*}

\begin{figure*}[t]
\includegraphics[width=\linewidth]{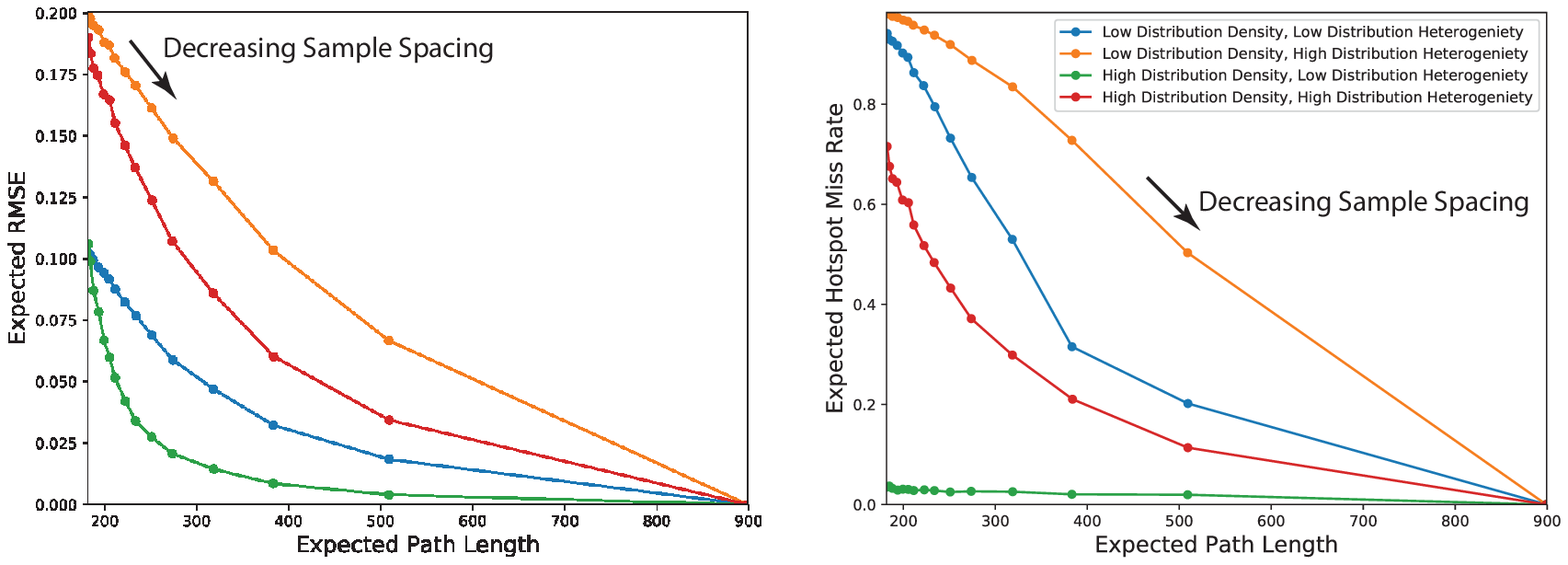}
\caption{The expected distribution estimation error plotted against the expected path length of the robot for varying sample spacing. Decreasing sample spacing results in lower estimation error, but longer robot paths.}
\label{fig:performanceVpathLength}
\end{figure*}

Although using a lower sample spacing leads to less recreation error and undetected hotspots, it also leads to a longer expected travel path and thus operation time. In general, the length of the path traveled by the robot will increase linearly with the area of the environment. For large environments, using a low sample spacing may not be practical, and the sample spacing needs to be chosen to balance information gain and operation costs. Figure \ref{fig:pathLengthVspacing} shows how the path length is expected to scale with different sample spacings given a fixed environment size of 30 by 30 and four different obstacle map types: low or high density (0.01 or 0.5) and low or high heterogeneity (0.1 or 0.75). Figure \ref{fig:performanceVpathLength} shows the trade-off between estimation accuracy and path length of the robot.

At high sample spacing, decreasing the sample spacing will result in a high information gain compared to the increase of robot path length. However, as sample spacing and recreation error decreases, reducing the sample spacing further will only result in a marginal gain in information compared to the increase of path length.

These results suggest that using boustrophedon decomposition as the robot's search algorithm is useful in two scenarios. First, it is useful when guarantees of low distribution recreation errors are needed and operational costs, such as travel distance or time, is not a concern. Here, very low sample spacing can be used to get nearly perfect distribution recreation. Second, it is useful when the distribution of the environment is expected to have high density and low heterogeneity. In these environments, larger sample spacings could be used and still result in low overall error and undetected hotspots.

\section{Improvements to Shorten Boustrophedon Path}
\label{sec:preprocessing}
\subsection{Thin Obstacle Preprocessing}
Section 4 shows that the boustrophedon decomposition algorithm has trouble with small fragmented cells in environments with many small obstacles. If the cell width is smaller than the desired path width, the robot's new path width becomes the cell width while rastering that cell. This results in longer path lengths since the robot needs to constantly travel and raster densely between the many small cells, even if the desired sample spacing is large. Furthermore, observe that as the sample spacing was increased, the path length started to become dominated by the distance it took to not only travel between all the small cells, but to also raster such cells. As a result, the path length did not decrease as the path width increased past a point, as we would expect, due to the presence of thin cells (Fig.~\ref{fig:pathLengthVspacing}).

\begin{figure*}[t]
\centering
\begin{subfigure}{0.4\textwidth} 
  \centering
  \includegraphics[trim=0 1 0 0, clip,width=\linewidth]{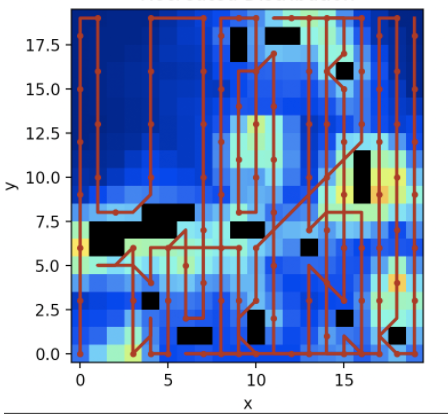}
  \caption{Robot path without preprocessing}
\end{subfigure}
\hfil
\begin{subfigure}{0.4\textwidth} 
  \centering
  \includegraphics[trim=0 1 0 5, clip,width=\linewidth]{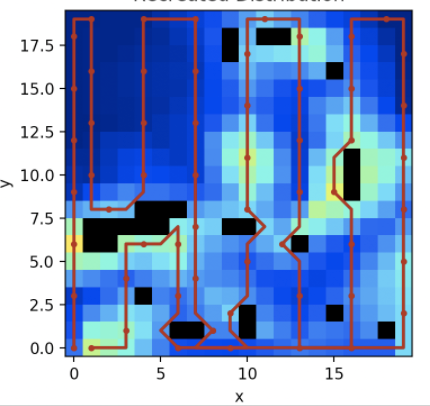}
  \caption{Robot path with preprocessing}
\end{subfigure}
\caption{Preprocessing an environment before using the boustrophedon cellular decomposition algorithm allows for less dense rastering and ultimately a shorter path length in an environment with many thin obstacles. In this example, the path width (sample spacing) was set to 4. Without preprocessing (a), most of the cells were quite narrow, resulting in areas of dense (width 1) rastering. This caused the robot to sample more frequently and travel further than desired. With preprocessing (b), the sample spacing is much more even, and as a result, the robot did not need to travel as far.}
\label{fig:preprocessingExample}
\end{figure*}

To address this issue, we implemented an environmental preprocessing algorithm. Since the robot can easily drive around smaller obstacles, they can be ignored during path planning. Our preprocessing algorithm removes obstacles that have a width less than the desired path width from the map before the boustrophedon algorithm splits the environment into cells. This prevents the boustrophedon algorithm from unnecessarily splitting thin cells, but also removes the safety guarantee associated with every cell. Each cell is no longer completely safely rasterable due to the possible thin obstacles inside. When the robot encounters ignored obstacles in a cell during operation, it simply goes around it and continues rastering. An example with and without this preprocesssing step is shown in Fig.~\ref{fig:preprocessingExample}.

\subsection{Results from Preprocessing Environment}
\begin{figure*}[t]
\centering
\begin{subfigure}{0.4\textwidth} 
  \centering
  \includegraphics[width=\linewidth]{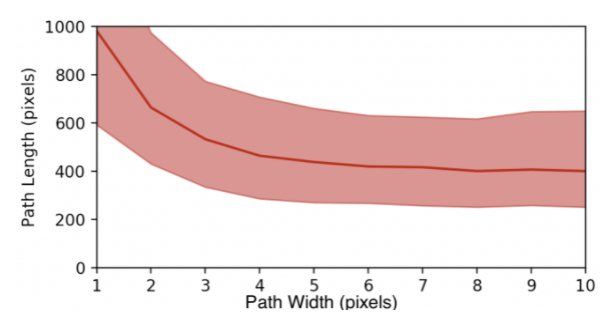}
  \caption{Robot path length and path width correlation without preprocessing}
\end{subfigure}
\hfil
\begin{subfigure}{0.4\textwidth} 
  \centering
  \includegraphics[width=\linewidth]{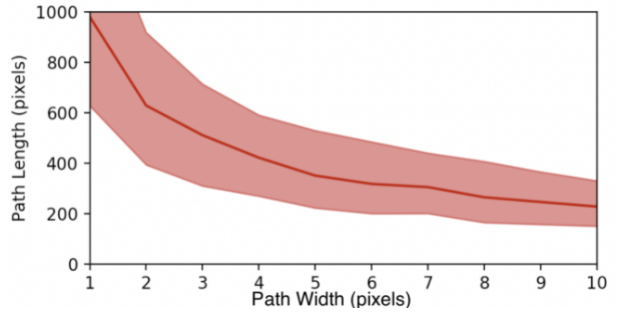}
  \caption{Robot path length and path width correlation with preprocessing}
\end{subfigure}
\caption{Preprocessing an environment before using the boustrophedon cellular decomposition algorithm allows for less dense rastering and ultimately a shorter path length in an environment with many thin obstacles. In this example, without preprocessing, the expected path length converged to around 400 at around a path width of 4 or higher. With preprocessing, the expected path length continued to decrease to around 200 at a path width of 10, a $50\%$ improvement.}
\label{fig:preprocessMCExample}
\end{figure*}

We also ran a Monte-Carlo simulation with and without preprocessing to determine the improvement with regards to path length. This Monte-Carlo simulation varied the path width and observed the expected path length for random environments with random distributions and varying obstacle densities.  The results are shown in Fig.~\ref{fig:preprocessMCExample}, with the average path length per sample spacing graphed as the red line, whereas the light red area represents the standard deviation.

Observe that the average path length with preprocessing trends downwards as path width increases, whereas the average path length without preprocessing plateaus, even as the path width increases. These results suggest that the use of preprocessing allows the boustrophedon decomposition algorithm to be used efficiently in regards to path length in larger environments with many thin obstacles.

\section{Conclusion}
\label{sec:Conclusion}
This work shows how the boustrophedon decomposition algorithm can be used to plan a path containing evenly spaced sampling locations in an obstacle-filled environment. Performance was evaluated in a Monte-Carlo simulation framework over a range of environments and metrics. This analysis shows that the boustrophedon decomposition algorithm can yield a perfect recreation of environment's distribution at the expense of higher path lengths and thus operation costs. This renders the algorithm appropriate for when information gain guarantees are valued over operational costs, or when the distribution is expected to be relatively uniform with larger hotspots (high density, low heterogeneity). This work also presents a modification to the boustrophedon algorithm which preprocesses the environment by removing obstacles thinner than the desired path width and routing around them during rastering, resulting in shorter path lengths with larger path widths.

Although the boustrophedon decomposition algorithm is works well in many environments and has preferable performance guarantees, many improvements can be made. First, the path length could be reduced by joining neighboring cells. Cellular decomposition partitions the environment into smaller obstacle free cells based on the shape of obstacles, but even if the environment is preprocessed many large obstacles or those with complex shapes will still be broken into many small cells (for example Cell 7 in Figure \ref{fig:example_decomposition}). This leads to longer robot paths since the robot needs to travel between and raster in each one of these small cells. Combining neighboring cells into larger cells can reduce the total amount of rastering and intercellular travel needed. The overall path length could also be reduced by optimizing the visit order of cells. As mentioned earlier, once the robot has finished rastering a cell, it greedily chooses to raster the closest uncovered cell next. More intelligent planners such as A* can be used to plan further ahead and minimize overall path length.

Improvements can also be made to the interpolation process. This work implements linear interpolation between sampled data for simplicity. Another common method of interpolating between spatial data is Gaussian process regression, known in geostatistics as kriging. Kriging models the distribution as a Gaussian or log-Gaussian process where the covariance between two data points are related to the spatial location of the two points \cite{cressie2015statistics}. This method of interpolation not only produces a smoother recreation of the distribution, but it also provides an uncertainty metric which can be useful in drawing conclusions about the underlying distribution.

One disadvantage of using boustrophedon decomposition is the fact that a long path lengths are needed to reduce RMSE and number of missed hotspots in low-density, low-heterogeneity environments, or environments with many small hotsposts. In such scenarios adaptive search algorithm techniques \cite{10.5555/1402383.1402392} could be used reduce uncertainty and number of undetected hotspots without resulting in an impractically long robot path. In some of these algorithms, the environment is modeled as a log-Gaussian process and sampling locations are intelligently chosen to reduce distribution recreation error. Although these methods generally require much longer computation times, faster optimization techniques such as cross-entropy method of optimization or reinforcement learning could be used to reduce computational overhead.

\section{Acknowledgements}
We thank Chevron CTC for partial funding support of this project.

\bibliography{references}
\end{document}